\documentclass[10pt,twocolumn,letterpaper]{article}

\usepackage{wacv}
\usepackage{times}
\usepackage{epsfig}
\usepackage{graphicx}
\usepackage{amsmath}
\usepackage{amssymb}
\usepackage{booktabs}

\usepackage{multirow} 
\usepackage{diagbox}

\usepackage{amsmath,amsfonts,bm}

\def\eqref#1{equation~\ref{#1}}

\def\1{\bm{1}}

\DeclareMathAlphabet{\mathsfit}{\encodingdefault}{\sfdefault}{m}{sl}
\SetMathAlphabet{\mathsfit}{bold}{\encodingdefault}{\sfdefault}{bx}{n}

\usepackage{setspace}

\usepackage{tabularx,colortbl}
\usepackage[dvipsnames]{xcolor}
\usepackage{tabulary}
\usepackage{etoolbox}
\usepackage{subcaption}
\usepackage{url}

\def\wacvPaperID{715} %

\wacvalgorithmstrack   %

\wacvfinalcopy %

\ifwacvfinal
\usepackage[breaklinks=true,bookmarks=false]{hyperref}
\else
\usepackage[pagebackref=true,breaklinks=true,colorlinks,bookmarks=false]{hyperref}
\fi

\begin{document}

\title{Out-of-distribution Detection via Frequency-regularized Generative Models}

\author{Mu Cai\\
Department of Computer Sciences\\
University of Wisconsin-Madison\\
{\tt\small mucai@cs.wisc.edu}
\and
Yixuan Li\\
Department of Computer Sciences\\
University of Wisconsin-Madison\\
{\tt\small sharonli@cs.wisc.edu}
}

\maketitle
\thispagestyle{empty}
\thispagestyle{plain}
\pagestyle{plain}

\newcommand{\method}{FRL}

\begin{abstract}

Modern deep generative models can assign high likelihood to inputs drawn from outside the training distribution, posing threats to models in open-world deployments. While much research attention has been placed on defining new test-time measures of OOD uncertainty, these methods do not fundamentally change how deep generative models are regularized and optimized in training. In particular, generative models are shown to overly rely on the background information to estimate the likelihood. To address the issue, we propose a novel frequency-regularized learning (\textbf{FRL}) framework for OOD detection, which incorporates high-frequency information into training and guides the model to focus on semantically relevant features. FRL effectively improves performance on a wide range of generative architectures, including variational auto-encoder, GLOW, and PixelCNN++. On a new large-scale evaluation task, FRL achieves the state-of-the-art performance, outperforming a strong baseline Likelihood Regret by {10.7}\% (AUROC) while achieving {147$\times$} faster inference speed. Extensive ablations show that FRL improves the OOD detection performance while preserving the image generation quality. Code is available at \url{https://github.com/mu-cai/FRL}.
    
\end{abstract}
\section{Introduction}

Modern deep generative models have achieved unprecedented success in known contexts for which they are trained, yet they do not necessarily know what they don’t know. In particular, Nalisnick \emph{et al.}~\cite{nalisnick2018deep} showed that generative models can produce abnormally high likelihood estimation for out-of-distribution (OOD) data---samples with semantics outside the training data distribution. Ideally, a model trained on MNIST should not produce a high likelihood score for an animal image, because the semantic is clearly different from hand-written digits. %

 This intriguing yet bewildering observation has triggered a plethora of literature to address the problem of OOD detection in generative modeling. Much of the prior work focused on defining more suitable test-time measures of OOD uncertainty, such as Likelihood Ratio~\cite{ren2019likelihood}, Input Complexity~\cite{serra2019input}, and Likelihood Regret~\cite{xiao2020likelihood}. Despite the improved performance, these methods do not fundamentally change how deep generative models are trained and optimized. 
 Arguably, continued research progress in OOD detection requires the improved design of learning methods, in addition to the inference-time statistical tests. This paper bridges this critical gap. 
 
In this paper, we propose a novel \emph{Frequency-Regularized Learning} framework for OOD detection (dubbed FRL). Our work is motivated by observations in Ren \emph{et al.}~\cite{ren2019likelihood}, which suggested that generative models' reliance on the background information undesirably leads to a high likelihood for OOD samples. Indeed, several recent studies~\cite{zhang2021understanding,nagarajan2021understanding,normalizeflow_fail} showed that current generative models overfit to the training data, particularly background pixels. To alleviate the issue, our key idea is to guide the model to pay more attention to the high-frequency information, which represents object contours and semantic details rather than the low-frequency image background. Shown in Figure~\ref{fig::example}, though OOD and in-distribution data have similar backgrounds, their high-frequency information  represents semantic feature differences well.  %
Our framework is grounded in classic signal processing~\cite{torrence1998practical, bracewell1986fourier, heideman1984gauss},
which demonstrated the efficacy of high-frequency components for capturing semantic content. 
In particular, \method{} adds the high-frequency component as an additional channel to the input image, which regularizes deep generative models to have less reliance on background information and thus improves test-time OOD detection. %

\method{} provides a general plug-and-play mechanism, which is applicable to common generative models, including Variational Auto-Encoder (VAE)~\cite{kingma2013auto,higgins2017beta}, GLOW~\cite{kingma2018glow}, and PixelCNN++~\cite{pixelcnn,salimans2017pixelcnn++}. Importantly, our method incurs minimal changes to the existing training architecture, and only requires modifying the number of the input channel. \method{} is straightforward and relatively simple to implement in practice. During the test time, we concatenate the original image with the high-frequency component, and estimate the OOD score on the concatenated input. 

\begin{figure*}[t] %
	\begin{center}
    \includegraphics[width=0.95\linewidth]{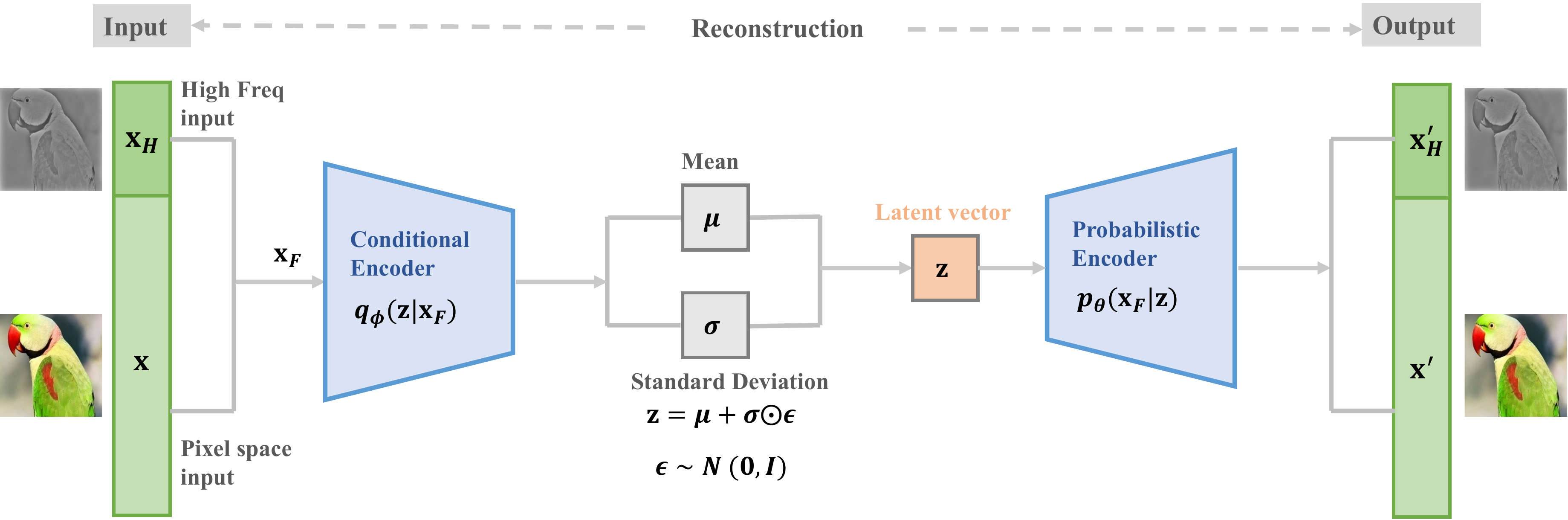} %
	\end{center}
	   \caption{Overview of the proposed \emph{frequency-regularized learning framework} (\textbf{\method{}}) for OOD detection. We exemplify using the VAE architecture. The key idea is to extract  the high-frequency information from the original image, and use that for regularizing the generative models. High frequency information captures the outline of objects as opposed to background.}
	    \vspace{-2ex}
	\label{fig:overall}
 \end{figure*}
 
We extensively evaluate our approach on various generative modeling architectures and datasets, where \method{} establishes state-of-the-art performance. On flow-based model GLOW, \method{} improves the performance on CIFAR-10 by 7.5$\%$ (AUROC), compared to the best baseline Input Complexity~\cite{serra2019input}. While prior literature has primarily evaluated performance on simple datasets such as CIFAR-10 and Fashion-MNIST, we further extend our evaluation to a  large-scale setting and test the limit of our approach. On CelebA dataset with higher resolution, \method{} outperforms a competitive method Likelihood Regret (LR)~\cite{xiao2020likelihood} by {10.7\%} in AUROC while achieving {147$\times$} faster inference speed on VAE. Unlike LR, \method{} alleviates the need for online optimization during inference time, a major bottleneck that prevents LR from performing real-time OOD detection. %
Our key contributions and results are summarized as follows:

\begin{itemize}
\vspace{-0.2cm}
    \item We propose a new frequency-regularized OOD detection framework \method{}, which regularizes the training of deep generative models by emphasizing high-frequency information. \method{} effectively improves performance on common generative modeling methods including VAE, GLOW, and PixelCNN++.
    \item We extensively evaluate \method{} on common benchmarks, along with a new large-scale evaluation task with high-resolution images. \method{} achieves the state-of-the-art performance, outperforming a strong baseline Likelihood Regret~\cite{xiao2020likelihood} by \textbf{10.7}\% (AUROC) while achieving \textbf{147$\times$} faster inference speed on CelebA. To the best of our knowledge, this is the first work that demonstrates the efficacy of generative-based OOD detection on datasets beyond the CIFAR benchmark.%
    \item We conduct extensive ablations to improve the understanding of the efficacy of our method, highlighting the importance of high-frequency. We show that \method{} improves the OOD detection performance while preserving the image generation quality. %
\end{itemize}

\section{Preliminaries}
\label{sec:pre}
 We consider the setting of unsupervised learning, where $\mathcal{X}$ denotes the input space. The training set $\mathcal{D} = \{\mathbf{x}_i\}_{i=1}^n$ is drawn \emph{i.i.d.} from in-distribution $P_{\mathcal{X}}$. This setting imposes weaker data assumption than discriminative-based OOD detection approaches, which require labeling information.

\paragraph{Out-of-distribution Detection} OOD detection can be viewed as a binary classification problem. At test time, the goal of OOD detection is to decide whether a sample $\mathbf{x} \in \mathcal{X}$ is from \emph{in-distribution} $P_{\mathcal{X}}$ (ID) or not (OOD). In practice, OOD is often defined by a distribution that simulates unknowns encountered during deployment time, such as samples from an irrelevant semantic (\eg MNIST vs. cat). The decision can be made via a thresholding mechanism:
\begin{align*}
\label{eq:threshold}
	G_{\lambda}(\mathbf{x})=\begin{cases} 
      \text{ID} & S(\mathbf{x})\le \lambda \\
      \text{OOD} & S(\mathbf{x}) > \lambda 
   \end{cases},
\end{align*}
where  samples with lower scores $S(\mathbf{x})$ are classified as ID and vice versa. The threshold $\lambda$ is typically chosen so that a high fraction of ID data (\eg 95\%) is correctly classified. A natural choice of scoring function is to directly  estimate  the negative log likelihood  of the  input using generative modeling, which we describe in the next.  %

\section{Method}

Our novel frequency-regularized out-of-distribution detection framework is illustrated in Figure~\ref{fig:overall}. 
In what follows, we first introduce the mechanism of extracting high-frequency information from an image (Section~\ref{sec::freq_info}). Our training object facilitates the preservation of frequency information during the generative modeling process (Section~\ref{sec::generative}).

\subsection{High Frequency Information}
\label{sec::freq_info}
Our work is motivated by prior work by Ren \emph{et al.}~\cite{ren2019likelihood}, which showed that generative models' reliance on the {background} undesirably leads to high likelihood estimation for OOD samples. For example, deep generative
models trained on CIFAR-10 can assign a higher
likelihood to OOD data from MNIST. To better understand the phenomenon, several recent studies~\cite{zhang2021understanding,nagarajan2021understanding,normalizeflow_fail} showed that current generative models overfit to the training data, especially the background pixels that are non-essential for determining the image semantics. In contrast, humans can distinguish MNIST images as OOD w.r.t. animal images, based on semantic information. 

\begin{figure}[t]
    \begin{subfigure}{.5\textwidth}
      \centering
      \includegraphics[width=0.9\linewidth]{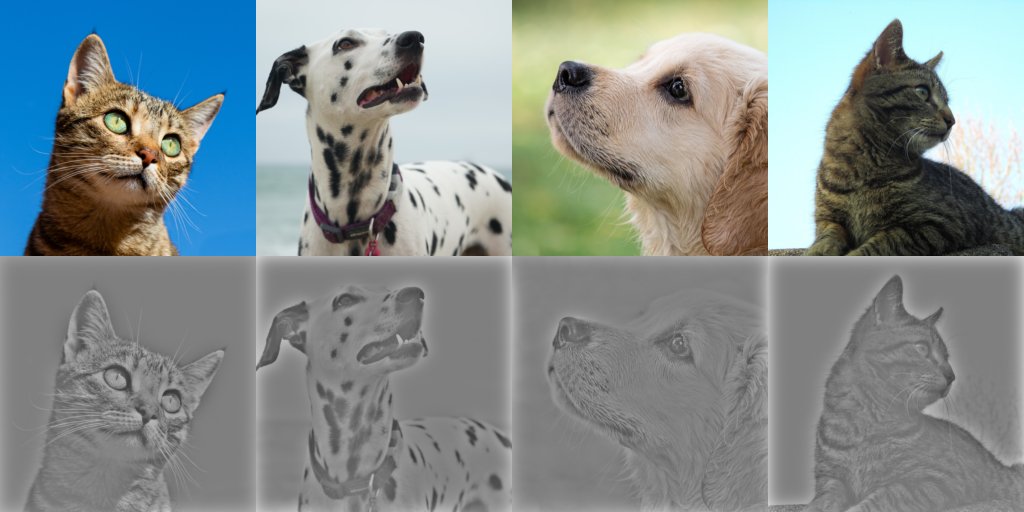}  
      \caption{In-distribution}
      \label{fig:sub-first1}
    \end{subfigure}
    \begin{subfigure}{.5\textwidth}
      \centering
      \includegraphics[width=0.9\linewidth]{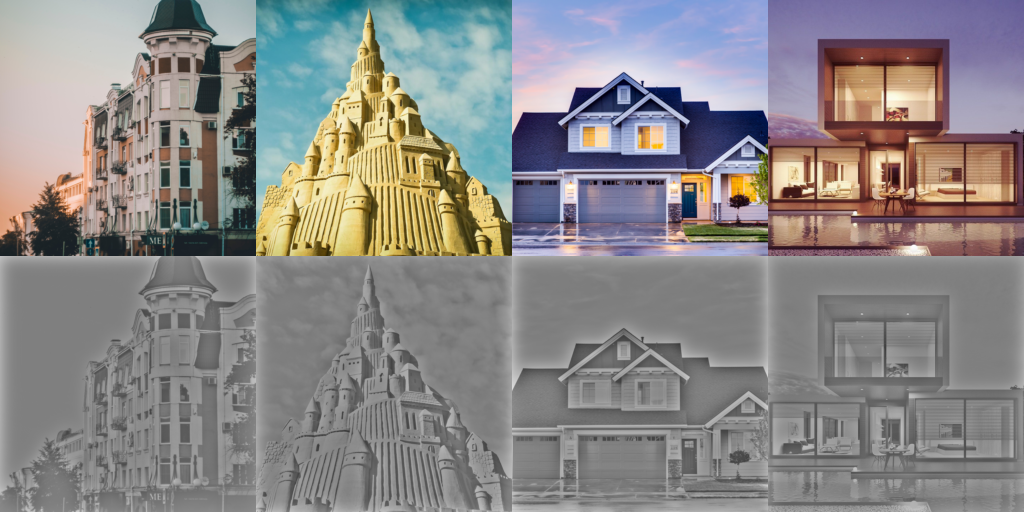}  
      \caption{Out-of-distribution}
      \label{fig:sub-second1}
    \end{subfigure}
    \caption{Visualization of the original RGB images and corresponding high-frequency features with similar backgrounds. In-distribution data  contains cats and dogs, which are different from out-of-distribution data buildings. } %
    \label{fig::example}
    \vspace{-3ex}
    \end{figure}

Motivated by this, the key idea of our framework is to exploit high-frequency information for enhancing generative-based OOD detection.
In particular, we alleviate the generative model's reliance on background information by guiding it to pay more attention to the high-frequency component of images. The high-frequency component is effective in capturing high-level semantic content, as established in classic signal processing literature~\cite{torrence1998practical, bracewell1986fourier, heideman1984gauss}. Compared to the color space image, high-frequency features can filter out the low-level background information and maintain the key semantic information, shown in Figure~\ref{fig::example}.

\begin{figure}[t]
	\begin{center}
    \includegraphics[width=\linewidth]{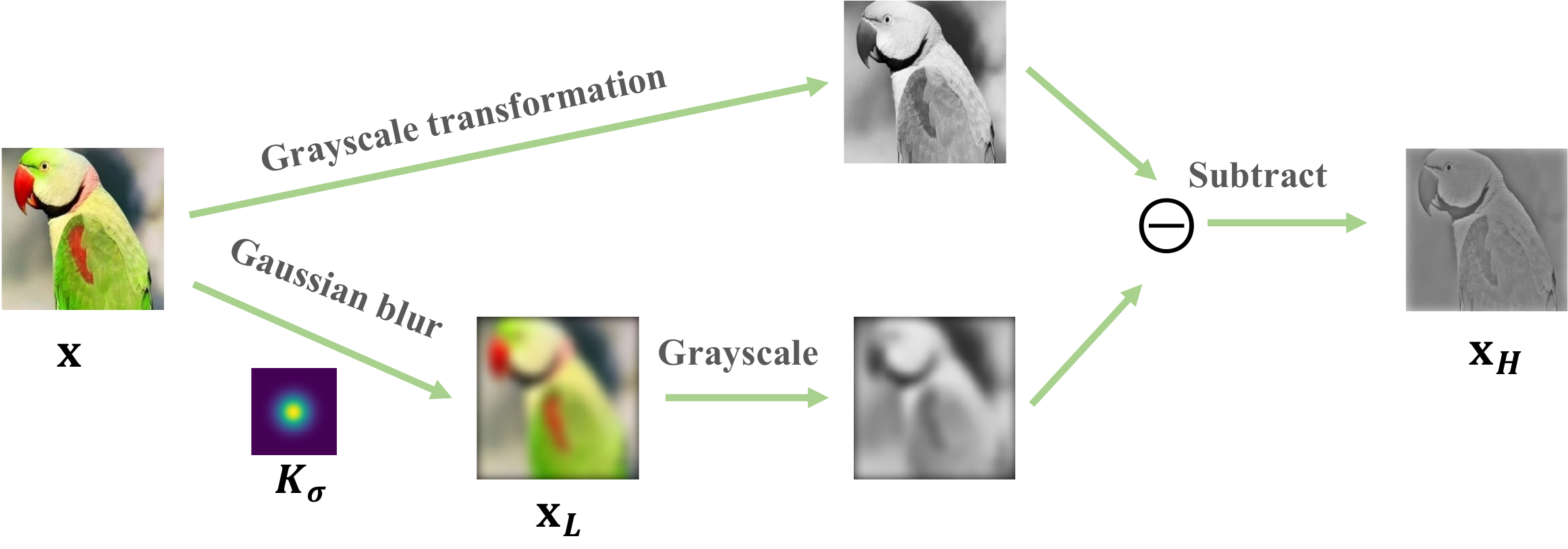} %
	\end{center}
	\vspace{-2ex}
	   \caption{Illustration on the process of extracting high frequency information. Gaussian blur is used to obtain the low frequency image. The high-frequency component is  the difference between the grayscale image and the blurred grayscale image.}
	    \vspace{-2ex}
	\label{fig:freq_create}
 \end{figure}

We now introduce details of how to transform the input image $\*\mathbf{x}\in \mathcal{X}$ into the high frequency counterpart  $\*\mathbf{x}_{H}$ . The overall procedure is shown in Figure~\ref{fig:freq_create}. Note that $\*\mathbf{x}_{H}$ has the same spatial dimension as $\*\mathbf{x}$. Specifically, we employ the Gaussian kernel $K_{\sigma}$:
\begin{equation}
	\begin{aligned}
		K_{\sigma}[m, n] &= \frac{1}{2 \pi \sigma^{2}} e^{-\frac{1}{2}\left(\frac{m^{2}+n^{2}}{\sigma^{2}}\right)},
	\end{aligned}
\end{equation}
where $[m, n]$ denotes the spatial location with respect to the center of an image batch, and $\sigma^{2}$ denotes the variance of the Gaussian function. Following~\cite{heideman1984gauss}, the variance is increased proportionally with the Gaussian kernel size. 
By  conducting convolution  on input $\*\mathbf{x}$ using  $K_{\sigma}$, we obtain the low frequency (\emph{blurred}) image $\mathbf{x}_L$:
\begin{equation}
	\begin{aligned}
		\mathbf{x}_L [i, j] &= \sum_{m=-\frac{k - 1}{2}}^{\frac{k - 1}{2}} \sum_{n=-\frac{k - 1}{2}}^{\frac{k - 1}{2}} K_{\sigma}[m, n] \cdot \mathbf{x}[i+m, j+n],
	\end{aligned}
\end{equation}
where $k$ denotes the kernel size, and $m, n$ denotes the index of an 2D Gaussian kernel, \textit{i.e.},   $m, n \in [-\frac{k - 1}{2}, \frac{k - 1}{2}]$.

To obtain the high-frequency image $\*\mathbf{x}_{H}$,
we first convert color images into grayscale images, and then subtract the low frequency information:
\begin{equation}
	\begin{aligned}
		\mathbf{x}_H &= { \texttt{rgb2gray} }(\mathbf{x})- [\texttt{rgb2gray}(\mathbf{x})]_{{L}},
	\end{aligned}
\end{equation}
where the $\texttt{rgb2gray}$ function converts the color image to the grayscale image. This operation removes the color and illumination information that is unrelated to the identity and structure. The resulting high-frequency image $\mathbf{x}_{{H}}$ contains the object outlines of the original image. We proceed by introducing the training objective that can leverage the high-frequency information.

\subsection{Generative Modeling with High-frequency Information}

To enforce the deep generative models to pay more attention to the high frequency, \ie, semantic features, we propose training deep generative models by adding high-frequency component to the input. In other words, we use input $\mathbf{x}_F = [\mathbf{x}, \mathbf{x}_H]$ via channel concatenation (see Figure~\ref{fig:overall}).
This way, the deep generative model is incentivized to learn the semantic information, because failure to recover the high-frequency component will incur a reconstruction loss.    
Our method incurs minimal changes to the architecture by only modifying the number of the input channel. 
In what follows, we consider three common generative modeling approaches including VAE, GLOW, and PixelCNN++. %

\label{sec::generative}

\vspace{-0.2cm}
\subsubsection{Variational Auto-Encoder (VAE)} 
VAE is a widely known approach for generative modeling~\cite{kingma2013auto,higgins2017beta}. The VAE consists of an encoder $q_{\phi}(\mathbf{z} \mid \mathbf{x}_F)$  and a decoder $p_{\theta}(\mathbf{x}_F \mid \mathbf{z})$ , as illustrated in Figure~\ref{fig:overall}.  Given a latent code $\mathbf{z}$ and its prior $p(\mathbf{z})$, the likelihood $p_{\theta}(\mathbf{x}_F)$ is modeled as:

\begin{equation}
    p_{\theta}(\mathbf{x}_F)=\int_{\mathcal{Z}} p_{\theta}(\mathbf{x}_F \mid \mathbf{z}) p(\mathbf{z}) \mathrm{d} \mathbf{z}.
    \end{equation}

During training, variational inference is utilized to minimize the evidence lower bound of the  log likelihood, which serves as a proxy for the true likelihood~\cite{kingma2013auto}:
\begin{footnotesize}
\begin{equation*}
    \begin{aligned}
    \small
    \log p_{\theta}(\mathbf{x}_F) & \geq \mathbb{E}_{q_{\phi}(\mathbf{z} \mid \mathbf{x}_F)}\left[\log p_{\theta}(\mathbf{x}_F \mid \mathbf{z})\right]-D_{\mathrm{KL}}\left[q_{\phi}(\mathbf{z} \mid \mathbf{x}_F) \| p(\mathbf{z})\right] \\
    & \triangleq \mathcal{L}(\mathbf{x}_F ; \theta, \phi),
    \end{aligned}
    \end{equation*}
\end{footnotesize}
where $q_{\phi}(\mathbf{z} \mid \mathbf{x}_F)$ is  the variational approximation to the true posterior distribution $p_{\theta}(\mathbf{z} \mid \mathbf{x}_F)$. 

During inference, it is not tractable to directly obtain the log likelihood $\log p_{\theta}(\mathbf{x}_F)$. Instead, the log likelihood is approximated by the importance weighted lower bound $\mathcal{L}_{K}(\mathbf{x}_F ; \theta, \phi)$:
\begin{footnotesize}
\begin{equation*}
\begin{aligned}
\log p_{\theta}(\mathbf{x}_F) & \geq \mathbb{E}_{\mathbf{z}^{1}, \ldots, \mathbf{z}^{K} \sim q_{\phi}(\mathbf{z}| \mathbf{x}_F)}\left[\log \frac{1}{K} \sum_{k=1}^{K} \frac{p_{\theta}\left(\mathbf{x}_F| \mathbf{z}^{k}\right) p\left(\mathbf{z}^{k}\right)}{q_{\phi}\left(\mathbf{z}^{k} | \mathbf{x}_F\right)}\right] \\ &\triangleq \mathcal{L}_{K}(\mathbf{x}_F ; \theta, \phi),
\end{aligned}
\end{equation*}
\end{footnotesize}
where $\mathbf{z}^{k}$ is a  Gaussian sample from the variational posterior $q_{\phi}(\mathbf{z} |\mathbf{x}_F)$.

\subsubsection{GLOW}
\label{glow}
GLOW~\cite{kingma2018glow} adopts the invertible networks~\cite{flowbase}, without using the encoder-decoder architecture. 
Specifically, $\mathbf{f}$  is composed of a sequence of transformations: $\mathbf{f}=\mathbf{f}_{1} \circ \mathbf{f}_{2} \circ \cdots \circ \mathbf{f}_{K}$, such that the relationship between $\mathbf{x}_F$ and latent code $\mathbf{z}$ can be modeled as:
\begin{equation}
    \mathbf{x}_F \stackrel{\mathbf{f}_{1}}{\longleftrightarrow} \mathbf{h}_{1} \stackrel{\mathbf{f}_{2}}{\longleftrightarrow} \mathbf{h}_{2} \cdots \stackrel{\mathbf{f}_{K}}{\longleftrightarrow} \mathbf{z},
\end{equation}
where $\mathbf{h}_{i}(i=1, \cdots, K-1)$ is the intermediate variable.   Such a sequence of invertible transformations is also called a (normalizing) flow. The latent variable $\mathbf{z}$ is generated as a descriptor for the input $\mathbf{x}_F$.  
 Then given a datapoint $\mathbf{x}_F$, the log probability density function  of
the model parameterized by $\theta$  can be written as:
\begin{equation}
\begin{aligned}
\log p_{\boldsymbol{\theta}}(\mathbf{x}_F) &=\log p_{\boldsymbol{\theta}}(\mathbf{z})+\log |\operatorname{det}(d \mathbf{z} / d \mathbf{x}_F)| \\
&=\log p_{\boldsymbol{\theta}}(\mathbf{z})+\sum_{i=1}^{K} \log \left|\operatorname{det}\left(d \mathbf{h}_{i} / d \mathbf{h}_{i-1}\right)\right|.
\end{aligned}
\end{equation}
In other words, the log-likelihood $\log p_{\boldsymbol{\theta}}(\mathbf{x}_F)$ is derived using the likelihood of $\mathbf{z}$  and invertible 1$\times$1 convolution modules.
The negative log-likelihood (bits per dimension) could be utilized for downstream tasks such as OOD detection.

\subsubsection{PixelCNN++}
\label{pixelcnn}
PixelCNN and PixelCNN++~\cite{pixelcnn,salimans2017pixelcnn++}  belong to the family of autoregressive models, which sequentially predict the output elements. Given an 2D image $\mathbf{x}_F$, PixelCNN++ generates the output image pixel by pixel. Therefore, the joint distribution of pixels over an image $\mathbf{x}_F$ can be decomposed into the
following product of conditional probabilities:
\begin{equation}
p(\mathbf{x}_F)=\prod_{i=1}^{n^{2}} p\left(x_{i} \mid x_{1}, \ldots, x_{i-1}\right),
\end{equation}
where $x_i$ is the pixel value in each location. The ordering of the pixel dependencies is in raster scan order: row by row and pixel by pixel within
every row. Therefore, each pixel depends on all the pixels above and to the left of it, and not on any of the other pixels. 

\subsection{OOD Detection Score}
\label{sec::freq_score}

While a straightforward idea is to employ the Negative Log Likelihood score, recent work~\cite{serra2019input} showed that it is more advantageous to subtract the Input Complexity, resulting in the following function:
\begin{equation}
    S(\mathbf{x})=-\log p_{\theta}(\mathbf{x})-L(\mathbf{x}),
    \end{equation}
where the complexity score $L(\mathbf{x})$ is represented by the code length derived from the data compressors~\cite{serra2019input}. This score function can also be interpreted from a likelihood-ratio test perspective.

Inspired by this, we employ the frequency-based log-likelihood term to derive a new scoring function $S_F(\mathbf{x})$ for OOD detection:
\begin{equation}
    S_F(\mathbf{x}) = -\log p_{\theta}(\mathbf{x}_F)-L(\mathbf{x}),
\label{eqn:adv_loss}
\end{equation}
where $S_F(\mathbf{x})$ captures frequency information for OOD detection. The image compression algorithm to represent the code length $L(\mathbf{x})$ is based on the Portable Network Graphics~(PNG)~\cite{png} format. %

\section{Experiments}

In this section, we first describe the experimental details (Section~\ref{sec::exp setup}), then we evaluate our approach \method{} on various generative modeling architectures and datasets (Section~\ref{sec:compare} \& Section~\ref{sec:large}). Further ablation studies are provided in Section~\ref{sec:Ablation}. Extensive experimental results show that \method{} not only preserves the image generation capability, but also enhances OOD detection.

\subsection{Experimental Details}
\label{sec::exp setup}  %
\paragraph{Common Benchmark} We use CIFAR-10~\cite{cifar10}, Fashion-MNIST~\cite{fmnist}  as the in-distribution datasets. For both datasets, we consider a total of 9 OOD datasets, which are all resized to 32$\times$32. The OOD datasets includes \texttt{SVHN}~\cite{svhn}, \texttt{LSUN}~\cite{lsun}, \texttt{MNIST}~\cite{mnist}, \texttt{KMNIST}~\cite{kmnist}, \texttt{Omniglot}~\cite{omniglot}, \texttt{NotMNIST}, \texttt{Noise}, and \texttt{Constant}. In CIFAR-10 evaluation, the OOD dataset also includes Fashion-MNIST, vice versa. %

\paragraph{Large-scale Evaluation Datasets} Furthermore, we also evaluate our method on large scale high-resolution dataset CelebA~\cite{celeba}. Following~\cite{Huang_2021_CVPR}, we adopt four high resolution OOD datasets: \texttt{iNaturalist}~\cite{Horn_2018_CVPR}, \texttt{Places}~\cite{xiao2010sun}, \texttt{SUN}~\cite{zhou2017places}, and \texttt{Textures}~\cite{Cimpoi_2014_CVPR}. All images are resized to $128\times128$.%

\paragraph{Training Details} We provide details for training on each of the architecture: VAE, GLOW and PixelCNN++.

(1) VAE is trained for 100 epochs for CIFAR-10 and Fashion-MNIST, and 110 epochs for CelebA.  %
The Gaussian kernel size is set to be 5, which we provide further ablation in Section~\ref{sec::ablation_kernel_size}. 
Following~\cite{xiao2020likelihood}, $p_{\theta}(\mathbf{x} \mid \mathbf{z})$  is chosen to be follow the 256-way discrete distribution. In image domain, this distribution corresponds to an 8-bit image on each pixel. 

 (2) GLOW is trained for 50 epochs with batch size 32 for both CIFAR-10 and Fashion-MNIST. The learning rate is $5 \times 10^{-4}$.  Following~\cite{kingma2018glow}, we adopt the invertible $1\times1$ convolutional (InvConv) layers in GLOW. 
 
 (3) PixelCNN++ is trained by 110 epochs with the learning rate $5 \times 10^{-4}$. There are overall  160 filters across the model. For the encoding part of the PixelCNN++, the model uses 3 residual
blocks consisting of 5 residual layers. %

\paragraph{Metric and Hardware} Following the literature~\cite{nalisnick2018deep,ren2019likelihood}, we primarily use AUROC as our evaluation metrics for OOD detection.  All experiments are conducted on NVIDIA GTX 2080Ti GPUs. %

\subsection{Evaluation on Common Benchmarks}
\label{sec:compare}

In this section, we evaluate our approach on the common benchmark, and compare it with competitive generative-based OOD detection methods. We consider the following baselines:  Negative Log Likelihood~(NLL)~\cite{nalisnick2018deep}, Likelihood Ratio (LRatio)~\cite{ren2019likelihood}, Input Complexity (IC)~\cite{serra2019input}, and two variants of Likelihood Regret (LR)~\cite{xiao2020likelihood}: LR(E) which optimizes the encoder, and LR(Z) which optimizes the latent variable. For fair comparison, all the baseline methods are trained and evaluated under consistent setting\footnote{Our implementation is based on the codebase: {\url{https://github.com/XavierXiao/Likelihood-Regret}}}. Our results reported are averaged across 5 independent runs.

\begin{table*}[t]
    \caption{AUROC values for OOD detection in  \textbf{GLOW} and \textbf{PixelCNN++}  when CIFAR-10 is the in-distribution dataset.}
    \label{tab::exp_cifar10_glow}
    \centering
    \small %
          \begin{tabular}{lccccp{0.0cm}ccccc}\toprule 
  Dataset & \multicolumn{4}{c}{GLOW} & \multicolumn{5}{c}{PixelCNN++}  \\
  \cline{2-5} \cline{7-10}
  
  & NLL &LRatio &IC &\method{}  & &NLL &LRatio &IC &\method{} \\
  & \cite{nalisnick2018deep} & \cite{ren2019likelihood} & \cite{serra2019input} & (ours) & & \cite{nalisnick2018deep} & \cite{ren2019likelihood} & \cite{serra2019input} & (ours) \\\midrule
SVHN &0.070 &0.161 &0.883 &0.915  $\pm$ 0.001 & &0.129 &0.949 &0.737 &0.831  $\pm$ 0.002\\
LSUN &0.890 &0.730 &0.213 &0.114 $\pm$ 0.002 & &0.852 &0.785 &0.640 &0.569  $\pm$ 0.004\\
MNIST &0.001 &0.003 &0.858 &0.961 $\pm$ 0.002  & &0.000 &0.092 &0.967 &0.999 $\pm$ 0.000 \\
FMNIST &0.007 &0.007 &0.712 &0.874 $\pm$ 0.003 & &0.003 &0.494 &0.907 &0.979  $\pm$ 0.001\\
KMNIST &0.007 &0.008 &0.380 &0.645 $\pm$ 0.002 & &0.002 &0.341 &0.826 &0.980  $\pm$ 0.001\\
Omniglot &0.000 &0.001 &0.955 &0.987 $\pm$ 0.000 & &0.000 &0.951 &0.989 &1.000  $\pm$ 0.000\\
NotMNIST &0.006 &0.009 &0.539 &0.720 $\pm$ 0.005 & &0.003 &0.718 &0.826 &0.979  $\pm$ 0.001\\
Noise &1.000 &0.426 &1.000 &1.000 $\pm$ 0.000 & &1.000 &1.000 &1.000 &1.000 $\pm$ 0.000 \\
Constant &0.010 &0.053 &1.000 &1.000 $\pm$ 0.000 & &0.042 &0.428 &1.000 &1.000  $\pm$ 0.000 \\
\cellcolor[HTML]{E3E3E3}Average &\cellcolor[HTML]{E3E3E3}0.221 &\cellcolor[HTML]{E3E3E3}0.155 &\cellcolor[HTML]{E3E3E3}0.727 &\cellcolor[HTML]{E3E3E3}\textbf{0.802$\pm$ 0.001 } &\cellcolor[HTML]{E3E3E3} &\cellcolor[HTML]{E3E3E3}0.226 &\cellcolor[HTML]{E3E3E3}0.640 &\cellcolor[HTML]{E3E3E3}0.877 &\cellcolor[HTML]{E3E3E3}\textbf{0.926 $\pm$ 0.001} \\\midrule
Num img/$s$ ($\uparrow$) &40.1 &	20.3 &	38.6 &	33.7 & & 20.0 & 		10.7 & 		19.3 & 	16.2 \\\midrule
 $T_{\text{inference}}(s)$ ($\downarrow$) & 0.025& 	0.049& 	0.026& 	0.030 & & 0.050 & 		0.093 & 	0.052 &	0.062\\
  \bottomrule
  \end{tabular}
  \vspace{-3ex}
\end{table*}

\paragraph{GLOW} The OOD detection results for CIFAR-10 based on GLOW model are shown in Table~\ref{tab::exp_cifar10_glow} (left). Flow-based models use invertible convolutions for estimating the likelihood, thus the parameters of the encoder and decoder are the same. Therefore, Likelihood Regret (LR) is not applicable here. \method{} establishes the state-of-the-art performance, outperforming the best baseline Input Complexity (IC) by {7.5}\% in AUROC. The comparison between our method and IC directly highlights the benefit of using high-frequency information for model regularization as well as inference. Note that Likelihood Ratio (LRatio) does not perform well on the GLOW model. We will contrast our method with LRatio on PixelCNN+ and VAE in the next, where LRatio is more effective. %

  \paragraph{PixelCNN++}
Table~\ref{tab::exp_cifar10_glow} (right) shows the OOD detection results using PixelCNN++~\cite{pixelcnn}, where \method{} outperforms the baselines. Among all the baselines, LRatio~\cite{ren2019likelihood} also attempts to mitigate the influence of background information using the Likelihood Ratio statistics. Compared to LRatio, \method{} displays an improvement of  \textbf{28.6}\% AUROC. This suggests that using frequency information for model regularization can be more effective in mitigating the influence of background. Note that due to the property of sequential prediction in autoregressive models, there are no latent variables and encoders. Hence, Likelihood Regret (LR) is not applicable here. %

\begin{table}[t]\centering
  \caption{AUROC values for OOD Detection in \textbf{VAE} when CIFAR-10 is the in-distribution dataset. }\label{tab::exp_cifar10}
  \resizebox{0.51\textwidth}{!}{
  \begin{tabular}{lccccccc}\toprule
  OOD Dataset &NLL &LRatio &LR(Z) &LR(E) &IC &\method{} \\
  & \cite{nalisnick2018deep} & \cite{ren2019likelihood} & \cite{xiao2020likelihood} & \cite{xiao2020likelihood} & \cite{serra2019input} & (ours)\\
  \midrule
SVHN &0.081 &0.050 &0.655 &0.959 &0.907 &0.854 $\pm$ 0.002 \\
LSUN &0.926 &0.952 &0.456 &0.403 &0.174 &0.449  $\pm$ 0.003\\
MNIST &0.000 &0.902 &0.759 &0.999 &0.984 &0.984  $\pm$ 0.000\\
FMNIST &0.033 &0.665 &0.732 &0.991 &0.992 &0.993  $\pm$ 0.001\\
KMNIST &0.011 &0.918 &0.755 &0.999 &0.981 &0.985  $\pm$ 0.000\\
Omniglot &0.000 &0.937 &0.637 &0.996 &0.988 &0.988 $\pm$  0.001\\
NotMNIST &0.030 &0.492 &0.737 &0.994 &0.988 &0.990 $\pm$  0.000\\
Noise &1.000 &1.000 &0.703 &0.999 &0.167 &0.925  $\pm$ 0.002\\
Constant &0.299 &0.353 &0.833 &0.995 &1.000 &1.000  $\pm$ 0.000\\
\cellcolor[HTML]{E3E3E3}Average &\cellcolor[HTML]{E3E3E3}0.264 &\cellcolor[HTML]{E3E3E3}0.697 &\cellcolor[HTML]{E3E3E3}0.696 &\cellcolor[HTML]{E3E3E3}\textbf{0.926} &\cellcolor[HTML]{E3E3E3}0.798 &\cellcolor[HTML]{E3E3E3}{0.906 $\pm$ 0.001} \\\midrule
Num img/$s$ ($\uparrow$)&240.8 &133.2 &1.3 &2.6 &238.8 &169.3 \\\midrule
 $T_{\text{inference}}(s)$ ($\downarrow$) &0.0042 &0.0075 &0.7438 &0.3783 &0.0042 &0.0059\\
  \bottomrule
  \end{tabular}
  }
  \vspace{-3ex}
  \end{table}
  
\paragraph{VAE}
Table~\ref{tab::exp_cifar10} shows the OOD detection results on CIFAR-10. 
We compare with a competitive baseline, Likelihood Regret (LR)~\cite{xiao2020likelihood}. Note that LR employs an online estimation, which incurs excessive inference time in its optimization. The best variant, namely LR(E), can only process 2.6 images per second. In contrast,
\method{} is computationally efficient (169.3 images per second measured by the inference speed), while achieving comparable performance. Moreover, compared to IC, the failure cases of ~\texttt{LSUN} and ~\texttt{Noise} are reduced significantly using our method \method{}. For example, when using \texttt{Noise} as OOD data, FRL improves the AUROC from 0.167 (IC) to 0.922. This is because the image code length is only the approximation
of the complexity score. We also show in appendix that FRL produces a more concentrated score distribution for ID data (green shade), benefiting the OOD detection .

When using Fashion-MNIST as the ID dataset,
\method{} achieves a strong performance across all OOD datasets, with an average AUROC score of 0.976. This can attribute to the simple structure of Fashion-MNIST data, where VAE in this case indeed estimates the likelihood more easily. The full results are in the appendix.%

\subsection{Evaluation on High Resolution Dataset}
\label{sec:large}
While prior literature has primarily evaluated OOD detection performance on simple datasets such as CIFAR-10 and Fashion-MNIST, we extend our evaluation to the large-scale setting and test the limit of our approach. In particular, we consider a model trained on CelebA~\cite{celeba}, a large-scale human face dataset. Table~\ref{tab::exp_celeba} shows the OOD detection results  on VAE. \method{} achieves the best performance of AUROC {0.984}, averaged over four diverse OOD test datasets. Noticeably, Likelihood Regret families perform poorly in this setting---the best variant LR(E) achieves an AUROC 0.877. This behavior is in sharp contrast with small datasets such as CIFAR-10 (\emph{c.f.} Section~\ref{sec:compare}), where generative models can overfit to the in-distribution data and as a result, allows Likelihood Regret to learn a large likelihood offset during online likelihood optimization. However, in a large-scale setting, generative models such as VAE may no longer overfit to the in-distribution data, and become less effective when using LR.  Our results demonstrate that \method{} can be flexibly used for both small and large scale datasets, and displays more stable performance than baseline approaches such as Likelihood Regret. To the best of our knowledge, this is the first work that demonstrates the efficacy of generative-based OOD detection on large-scale datasets (relative to CIFAR).

  \begin{table}[htbp]\centering
    \caption{AUROC values for OOD Detection  in VAE  when CelebA is the in-distribution dataset. }\label{tab::exp_celeba}
    \small %
    \resizebox{0.48\textwidth}{!}{
    \begin{tabular}{cccccccc}\toprule
    OOD Dataset&NLL &LRatio &LR(Z) &LR(E) &IC &\method{} \\
    & \cite{nalisnick2018deep} & \cite{ren2019likelihood} & \cite{xiao2020likelihood} & \cite{xiao2020likelihood} & \cite{serra2019input} & (ours) \\
    \midrule
iNaturalist &0.993 &0.969 &0.415 &0.808 &0.955 &0.995 $\pm$ 0.000 \\
Places &0.933 &0.847 &0.744 &0.928 &0.976 &0.991 $\pm$ 0.000 \\
SUN &0.945 &0.884 &0.726 &0.929 &0.959 &0.987 $\pm$ 0.001 \\
Textures &0.938 &0.891 &0.465 &0.842 &0.918 &0.965 $\pm$ 0.001 \\
    \cellcolor[HTML]{E3E3E3}Average &\cellcolor[HTML]{E3E3E3}0.952  &\cellcolor[HTML]{E3E3E3}0.898 &\cellcolor[HTML]{E3E3E3}0.588 &\cellcolor[HTML]{E3E3E3}0.877 &\cellcolor[HTML]{E3E3E3}0.952 &\cellcolor[HTML]{E3E3E3}\textbf{0.984 $\pm$ 0.000 } \\\midrule
Num img/$s$ ($\uparrow$) &99.3 &6.3 &0.6 &0.3 &44.3 &41.2 \\ \midrule
$T_{\text{inference}}(s)$ ($\downarrow$)  &0.0101 &0.1586 &1.5643 &3.5843 &0.0226 &0.0243 \\
    \bottomrule
    \end{tabular}
    }
    \vspace{-2ex}
    \end{table}

\subsection{Ablation and Further Analysis}
\label{sec:Ablation}

\paragraph{High-frequency information is critical.}
\label{sec:exp_recon}

\begin{figure}[htbp]
   \centering
    \begin{subfigure}[t]{0.45\textwidth}
      \centering
      \includegraphics[width=0.9\linewidth]{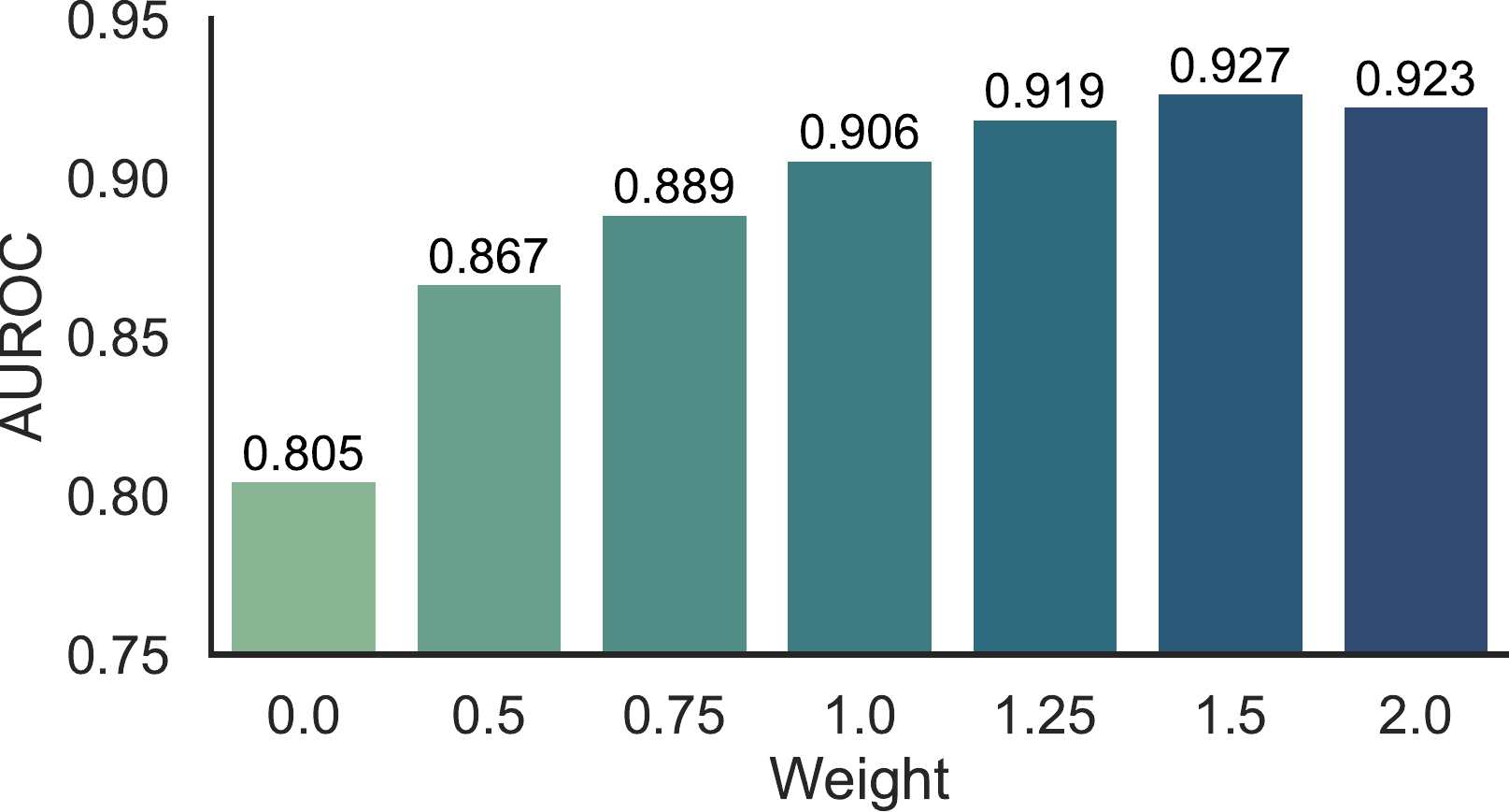}  
      \vspace{-1ex}
      \caption{Ablation on high-frequency importance}
      \label{fig:sub-second}
    \end{subfigure}
     \begin{subfigure}[t]{0.45\textwidth}
      \centering
      
\includegraphics[width=0.9\linewidth]{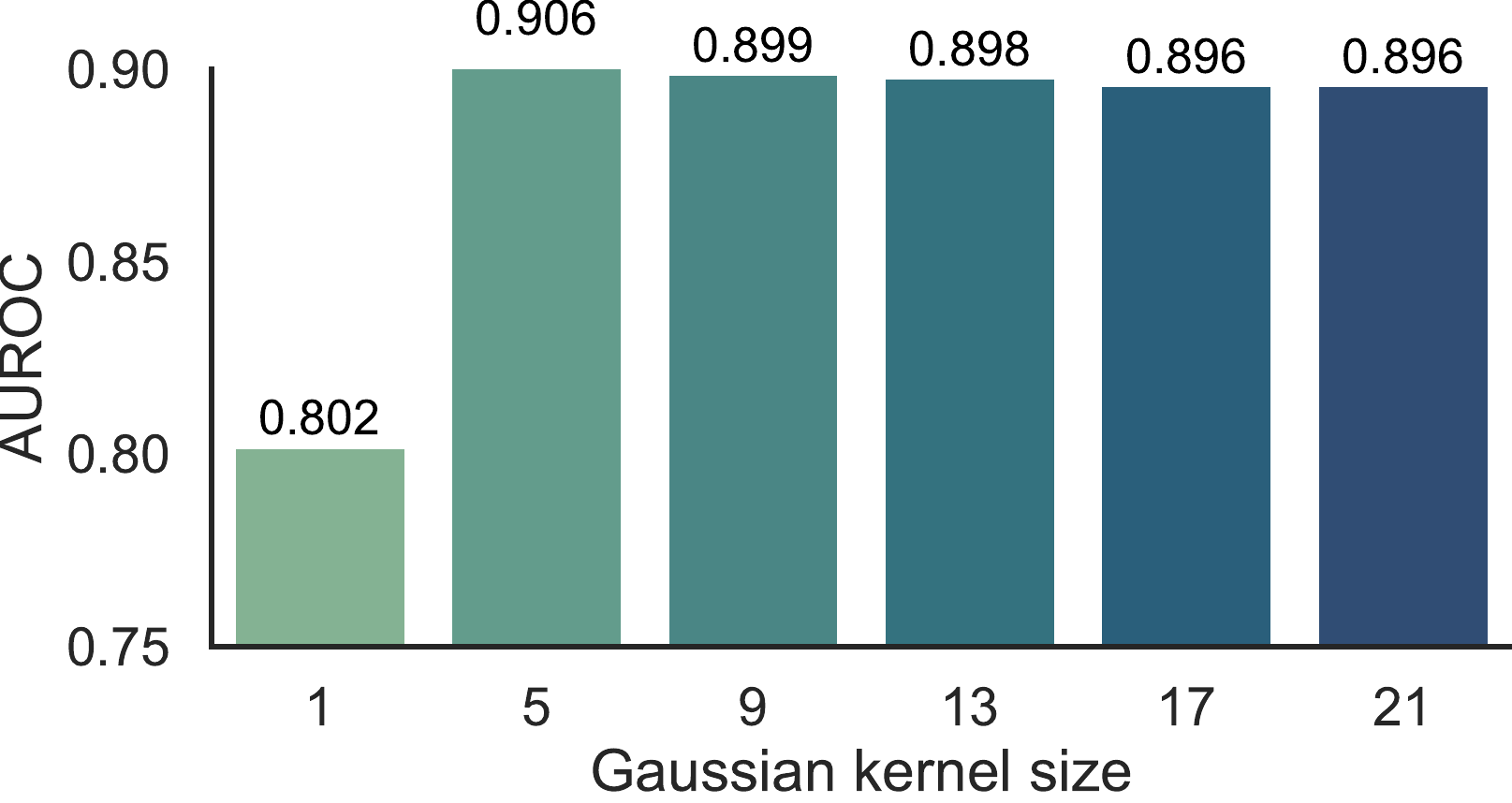}  
\vspace{-1ex}
      \caption{Ablation on Gaussian kernel sizes}
      \label{fig:sub-second5}
    \end{subfigure}
    \caption{Ablation on high-frequency importance and Gaussian kernel sizes in VAE. CIFAR-10 is the in-distribution data. Results are averaged over all OOD test datasets.} %
    \vspace{-1ex}
    \label{fig:kernel}
    \end{figure}

We demonstrate that frequency information is critical in inference-time OOD detection. Here we use the VAE framework to exemplify this. Recall that VAE has three components in the importance weighted lower bound during evaluation: $p_{\theta}\left(\mathbf{x}_F \mid \mathbf{z}^{k}\right)$, $p\left(\mathbf{z}^{k}\right)$, $q_{\phi}\left(\mathbf{z}^{k} \mid \mathbf{x_F}\right)$. In particular,  $p_{\theta}\left(\mathbf{x}_F \mid \mathbf{z}^{k}\right)$ denotes the likelihood of reconstruction for the input $\mathbf{x}_F = [\mathbf{x}, \mathbf{x}_H]$. Since the reconstruction is operated pixel-wise, we can split $p_{\theta}\left(\mathbf{x}_F \mid \mathbf{z}^{k}\right)$ into two parts:
$p_{\theta}\left(\mathbf{x} \mid \mathbf{z}^{k}\right)$ and  $p_{\theta}\left(\mathbf{x}_H \mid \mathbf{z}^{k}\right)$, which represent the contribution from the original image and high-frequency information respectively. To isolate the effect of high-frequency information, we vary the weight of $p_{\theta}\left(\mathbf{x}_H \mid \mathbf{z}^{k}\right)$ from 0 to 2  in the log likelihood form in inference time. The average AUROC is shown in Figure~\ref{fig:kernel} (up). 
When the high-frequency reconstruction part is completely removed (corresponding to 0), OOD detection performance degrades significantly. This becomes equivalent to the IC baseline. When we further increase the high-frequency weight to {1.5}, \method{} achieves an AUROC value of {0.927}, which matches the performance of LR(E) using costly online optimization. 

To further validate, when removing high-frequency information from the source image,  AUROC would decrease from 0.906 to 0.834 for VAE when CIFAR-10 serves as the in-distribution data. This ablation study again demonstrates the importance of high frequency reconstruction in generative-based OOD detection. Another example is that  for Fashion MNIST in VAE,  the mean square reconstruction error for high frequency information decreases from $2.12 \times 10^{-3}$ to $1.34 \times 10^{-3}$ by adopting \method{}, where our methods achieves strong performance
across all OOD datasets with average AUROC of 0.976.

Note that there is no reconstruction process in GLOW and PixelCNN++, where they directly produce the likelihood. Therefore we focus on the VAE  since the disentanglement of the likelihood from the original image and high-frequency image is not meaningful for GLOW and PixelCNN++. %

\paragraph{Ablation on the Gaussian kernel sizes.}
\label{sec::ablation_kernel_size}
Recall that the Gaussian kernel is adopted to induce the \emph{cf} Section~\ref{sec::freq_info}). The kernel size controls the tradeoff between the high-level structures and texture details. We train  VAEs with different Gaussian kernel sizes on CIFAR-10, and evaluate the OOD detection performance respectively. The average AUROC across all OOD datasets is shown in Figure~\ref{fig:kernel} (down). Results show that \method{} is not sensitive to the choice of kernel size when it is sufficiently large.  %

\paragraph{FRL is compatible with alternative high-frequency representations.} In \method{}, the Gaussian kernel aims at exploiting high-frequency information to enhance OOD detection performance. In addition to the Gaussian kernel, we show that other mechanisms for extracting high-frequency information can also be used in our framework, such as  Fourier transformation with FFT~\cite{Cai_2021_ICCV} and Wavelet transformation using Haar wavelets~\cite{Yoo_2019_ICCV}. Table~\ref{tab:rebuttal_high_freq_rep} shows that the performance of all three methods exploiting high-frequency information surpasses the baseline Input Complexity by a large margin. On GLOW, the Wavelet transformation shows slight improvement over the Gaussian kernel. %

\begin{table}[t]\centering
\caption{AUROC values for OOD detection under three high frequency representation forms in VAE, GLOW, and PixelCNN++ when CIFAR-10 is the in-distribution dataset, here results are averaged upon all OOD datasets.}\label{tab:rebuttal_high_freq_rep}
\resizebox{0.48\textwidth}{!}{
\begin{tabular}{lcccc}\toprule
High-frequency  form &VAE &GLOW &PixelCNN++ \\\midrule %
None (Input Complexity) &0.798 &0.727 &0.877 \\\midrule
Gaussian kernel &\textbf{0.906} &0.802 &\textbf{0.920} \\
Fourier transformation &0.860 &0.809 &0.893 \\
Wavelet transformation &0.894 &\textbf{0.826} &0.905 \\
\bottomrule
\end{tabular}
}
  \vspace{-2ex}
\end{table}

\paragraph{\method{} achieves strong results with efficient inference.}
We show the number of images processed per second under different approaches for VAE in Table~\ref{tab::exp_cifar10} when CIFAR-10 serves as the ID dataset. Specifically,
\method{} can process 169.3 images per second, while the current best generative approach LR(E) can only process 1.3 images per second, which is far from the real-time inference requirement.
Furthermore, LR(E) can be much slower when deployed into real-world datasets with higher resolution. For example, as we show in Section~\ref{sec:large}, LR(E) processes 0.3 images per second on a larger dataset CelebA, which is \textbf{147$\times$} slower than our method.

\section{Related Work}

\paragraph{Out-of-distribution Detection}
Machine learning models commonly make the closed-world assumption that the training and testing data distributions match. However, this assumption rarely holds in the real world, where a model can encounter out-of-distribution data. OOD detection is thus critical to enabling safe model deployment. The idea of OOD detection is to reject samples from the unfamiliar distribution, and handle those with caution. A comprehensive survey on OOD detection is provided in~\cite{yang2021oodsurvey}. Existing OOD approaches can be categorized broadly into generative-based and discriminative-based approaches. The key difference between the two is the presence or absence of label information. Below we review literature in the two categories separately.  

\paragraph{Generative-based OOD Detection}
Generative modeling aims at estimating the likelihood of the given input~\cite{kingma2013auto}. Generative models can be roughly categorized into three types: auto-encoders~\cite{kingma2013auto}, flow-based models~\cite{flowbase}, and autoregressive~\cite{pixelcnn,salimans2017pixelcnn++} models. Autoencoder-based models~\cite{an2015variational}  aim at reconstructing the input using the encoder-decoder architecture. Flow-based models~\cite{flowbase} such as GLOW~\cite{flowbase}  adopt the invertible network architectures to derive the likelihood from the latent variable. Autoregressive models like PixeCNN~\cite{pixelcnn}  and PixelCNN++~\cite{salimans2017pixelcnn++} sequentially predict the likelihood for each element, and optimize the joint likelihood over all elements. The generative model is a natural choice for OOD detection. Intuitively, the likelihood of the in-distribution samples should be higher than that of the out-of-distribution %
samples. However, Nalisnick \emph{et al.}~\cite{nalisnick2018deep} revealed that deep generative models trained on CIFAR-10 unexpectedly assign a higher likelihood to certain OOD datasets such as MNIST. 
Subsequent works propose different test-time measurements such as Likelihood Ratio~\cite{ren2019likelihood}, Input Complexity~\cite{serra2019input}, and Likelihood Regret~\cite{xiao2020likelihood}. To better understand the phenomenon, several recent studies~\cite{zhang2021understanding,nagarajan2021understanding,normalizeflow_fail} showed that current generative models overfit to the training set, especially when the structures of the images are simple. 
In this paper, we propose to use high-frequency information to regularize the training of deep generative models, which in turn improves the test-time OOD detection substantially.

\paragraph{Discriminative-based OOD Detection}
A parallel line of the OOD detection approach relies on discriminative-based models, which utilize label information. The phenomenon of neural networks' overconfidence in out-of-distribution data is first revealed by Nguyen \emph{et al.}~\cite{nguyen2015deep}, which makes softmax probability confidence unable to effectively distinguish between the ID and OOD data. 
Subsequent works attempted to improve the OOD uncertainty estimation by using OpenMax score~\cite{bendale2016towards}, deep ensembles~\cite{lakshminarayanan2017simple}, ODIN score~\cite{godin2020CVPR,liang2017enhancing}, and the energy score~\cite{liu2020energy}.
Besides model-based OOD detection, several works explored using features information to discriminative ID vs OOD data. For example, Mahalanobis distance~\cite{lee2018simple} uses the multivariate Gaussian model to estimate the feature space, then apply the distance score in the feature space to detect OOD samples. 
However, a drawback of discriminative-based OOD detection methods is that they don't directly model the likelihood of the input $p(\mathbf{x})$. In other words, the OOD scores can be less interpretable from a likelihood perspective.

\paragraph{Frequency Analysis in Deep Learning}
Frequency domain analysis is widely used in traditional image processing~\cite{heideman1984gauss,cooley1987re,van1992computational,johnson2006modified,gentleman1966fast}. The key idea of frequency analysis is to map the pixels from the Euclidean space to a frequency space, based on the changing speed in the spatial domain. Several works tried to bridge the connection between deep learning and frequency analysis~\cite{xu2020learning,chen2019drop,Xu_2020,10.1007/978-3-030-36708-4_22,NIPS2016_36366388,NEURIPS2018_a9a6653e}. 
Wang~\etal~\cite{wang2020high} found that the high-frequency components are useful in explaining the generalization of neural networks.  Durall~\etal~\cite{durall2020watch} observed that the images generated by GANs are heavily distorted in high-frequency parts, where they introduced a spectral regularization term to the loss function to alleviate this problem. 
Recent work FDIT~\cite{Cai_2021_ICCV} indicated that the high-frequency information better enhances the identity preserving in the image generation process. 
To the best of our knowledge, no prior work has explored using frequency-domain analysis for the out-of-distribution detection task. In this work, we propose a novel frequency regularized {OOD detection} framework, which demonstrates its superiority in terms of both OOD detection performance and computational efficiency.

\section{Conclusion}
In this work, we propose a novel frequency-regularized learning ({\method{}}) framework for out-of-distribution detection, which jointly estimates the likelihood of the pixel-space input and high-frequency information. Unlike existing generative modeling approaches, \method{} guides the model to focus on semantically relevant features during training, where high-frequency information helps regularize the model in training. \method{} can be flexibly used for common generative model architectures including VAE, GLOW, and PixelCNN++. Experiments on both common benchmark and a large-scale evaluation show that \method{} effectively improves the OOD detection performance while preserving the generative capability. Extensive ablation studies further validate the effectiveness of our approach both qualitatively and quantitatively. 
We hope our work will increase the
attention toward a broader view of frequency-based approaches for uncertainty estimation.

{\small
\bibliographystyle{ieee_fullname}
\bibliography{egbib}
}

\clearpage

\appendix
\onecolumn
\begin{center}
      {\large \bf {Supplementary Material} \par}
     
      \vskip .5em
      \vspace*{12pt}
\end{center}

\section{OOD Detection Results on Fashion-MNIST}

Table~\ref{tab::exp_vae_fmnist} shows the OOD detection results on Fashion MNIST in VAE. Our method \method{}
achieves a strong performance across all OOD datasets,
with an average AUROC score of 0.976. Among the challenging OOD datasets for Input Complexity (IC)~\cite{serra2019input}, such as  \texttt{LSUN} and \texttt{Noise}, \method{} achieves nearly optimal OOD detection performance. 

\begin{table}[htbp]\centering
  \caption{AUROC values for OOD Detection in VAE when Fashion-MNIST is the in-distribution dataset. }\label{tab::exp_vae_fmnist}
  \scriptsize
  \resizebox{0.65\textwidth}{!}{
  \begin{tabular}{lrrrrrrr}\toprule
  OOD Dataset &NLL &LRatio &LR(Z) &LR(E) &IC &\method{} \\
  & \cite{nalisnick2018deep} & \cite{ren2019likelihood} & \cite{xiao2020likelihood} & \cite{xiao2020likelihood} & \cite{serra2019input} & (ours)\\
  \midrule
SVHN &0.998 &0.546 &0.795 &1.000 &0.999 &1.000 \\
LSUN &1.000 &0.991 &0.691 &0.995 &0.508 &1.000 \\
CIFAR-10 &1.000 &0.929 &0.751 &0.999 &0.905 &1.000 \\
MNIST &0.165 &0.864 &0.589 &0.961 &0.932 &0.909 \\
KMNIST &0.630 &0.967 &0.620 &0.992 &0.629 &0.887 \\
Omniglot &1.000 &1.000 &0.725 &1.000 &1.000 &1.000 \\
NotMNIST &0.979 &0.965 &0.721 &1.000 &0.909 &0.988 \\
Noise &1.000 &1.000 &0.603 &0.999 &0.490 &1.000 \\
Constant &0.938 &0.416 &0.761 &0.998 &1.000 &1.000 \\
\cellcolor[HTML]{E3E3E3}Average &\cellcolor[HTML]{E3E3E3}0.857 &\cellcolor[HTML]{E3E3E3}0.853 &\cellcolor[HTML]{E3E3E3}0.695 &\cellcolor[HTML]{E3E3E3}{0.994} &\cellcolor[HTML]{E3E3E3}0.819 &\cellcolor[HTML]{E3E3E3}{\textbf{0.976}} \\\midrule
Num img/$s$ ($\uparrow$)&557.0 &284.4 &2.6 &1.3 &360.0 &262.0 \\\midrule
 $T_{\text{inference}}(s)$ ($\downarrow$) & 0.0018 &0.0035 &0.3779 &0.7419 &0.0028 &0.0038 \\
  \bottomrule
  \end{tabular}
  }
  \end{table}

  \section{Diagnosing the failure cases of prior approaches}
\label{sec:exp_problem}
\begin{figure}[htbp]
    \begin{subfigure}{.45\textwidth}
      \centering
      \includegraphics[width=\linewidth]{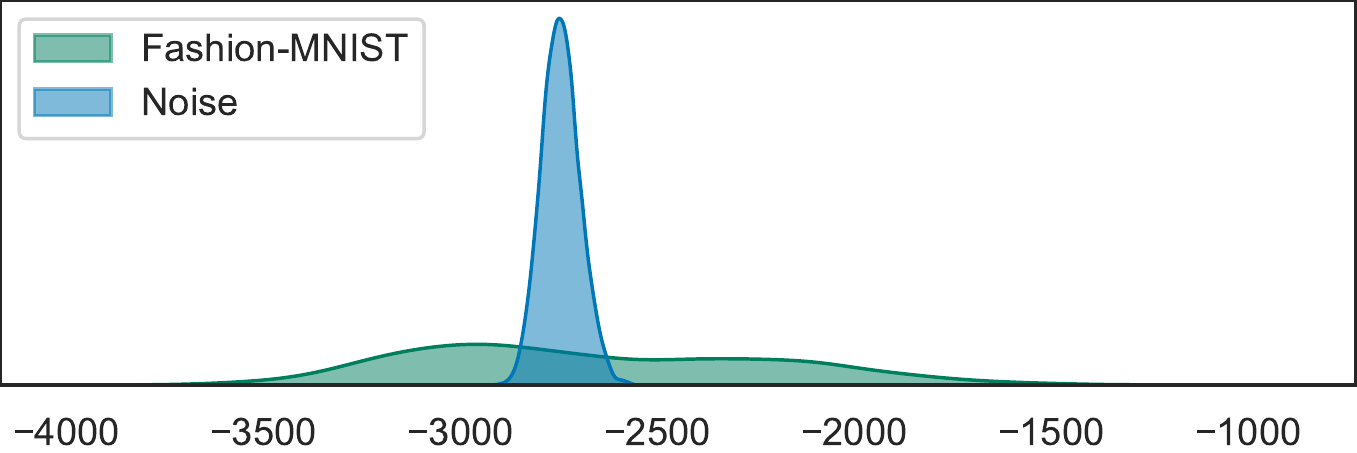}  
      \caption{IC}
      \label{fig:sub-first}
    \end{subfigure}
    \begin{subfigure}{.45\textwidth}
      \centering
      \includegraphics[width=\linewidth]{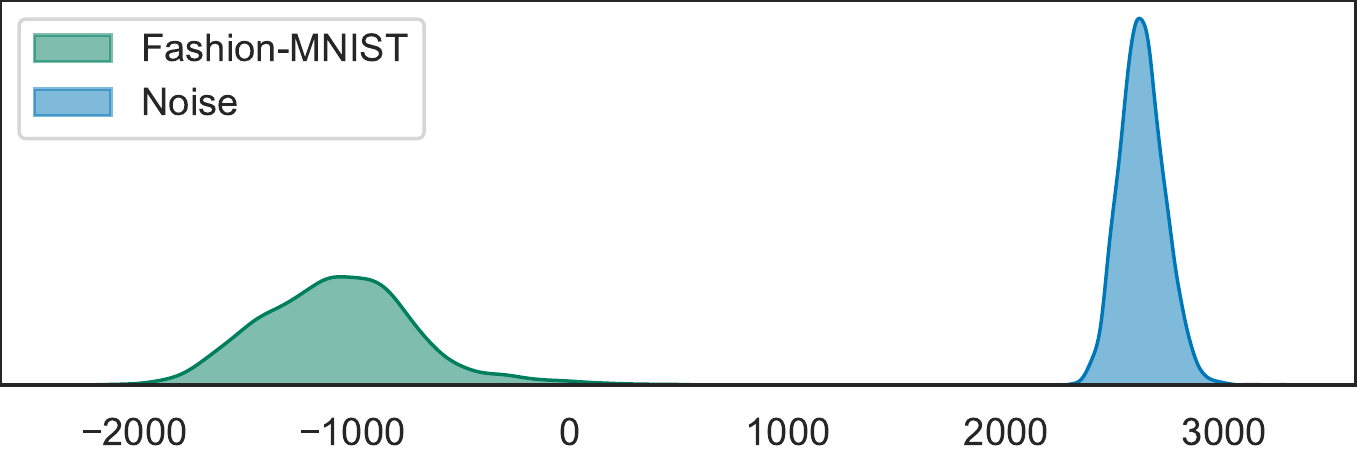}  
      \caption{\method{}}
      \label{fig:sub-second3}
    \end{subfigure}
    \caption{The  distribution of OOD scores  $S_F(\mathbf{x})$ for Fashion-MNIST and \texttt{Noise} dataset when Fashion-MNIST serves as the in-distribution dataset. Two panels denote the distributions under IC and \method{}, respectively.} %
    \label{fig::noise_dist}
    \end{figure}

Though the complexity score (IC)  mitigates many failure cases compared to directly employing NLL, IC struggles with a few special failure cases. In particular, we show the score distribution when the \texttt{Noise} dataset serves as the OOD dataset for VAE trained on Fashion-MNIST (ID). Shown in Figure~\ref{fig::noise_dist}, 
the scores for \texttt{Noise} lie in the middle of the scores for Fashion-MNIST, which is undesirable. This is because the image code length is only the approximation of the complexity score. In contrast, \method{} enables effective model regularization, which better distinguishes \texttt{Noise} and Fashion-MNIST data. Moreover, we also notice that \method{} produces a more concentrated score distribution for ID data (green shade), benefiting the OOD detection. 

\section{Ablations on Gaussian Kernel Sizes }  
 
\begin{figure}[htbp]
   
    \begin{subfigure}[t]{0.45\textwidth}
      \centering
      \includegraphics[width=\linewidth]{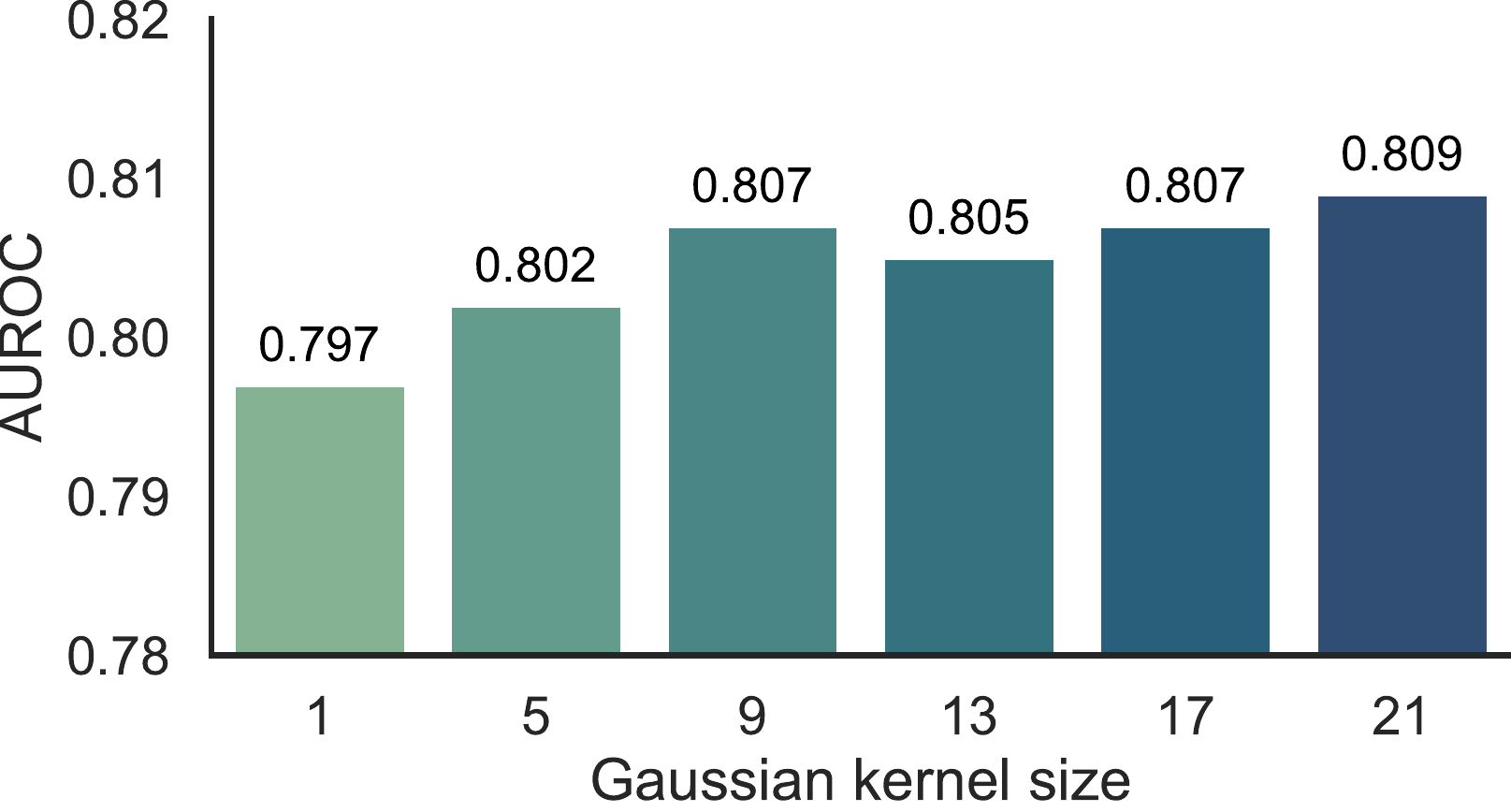}  
      \caption{GLOW}
      \label{fig:sub-first2}
    \end{subfigure}
     \begin{subfigure}[t]{0.45\textwidth}
      \centering
\includegraphics[width=\linewidth]{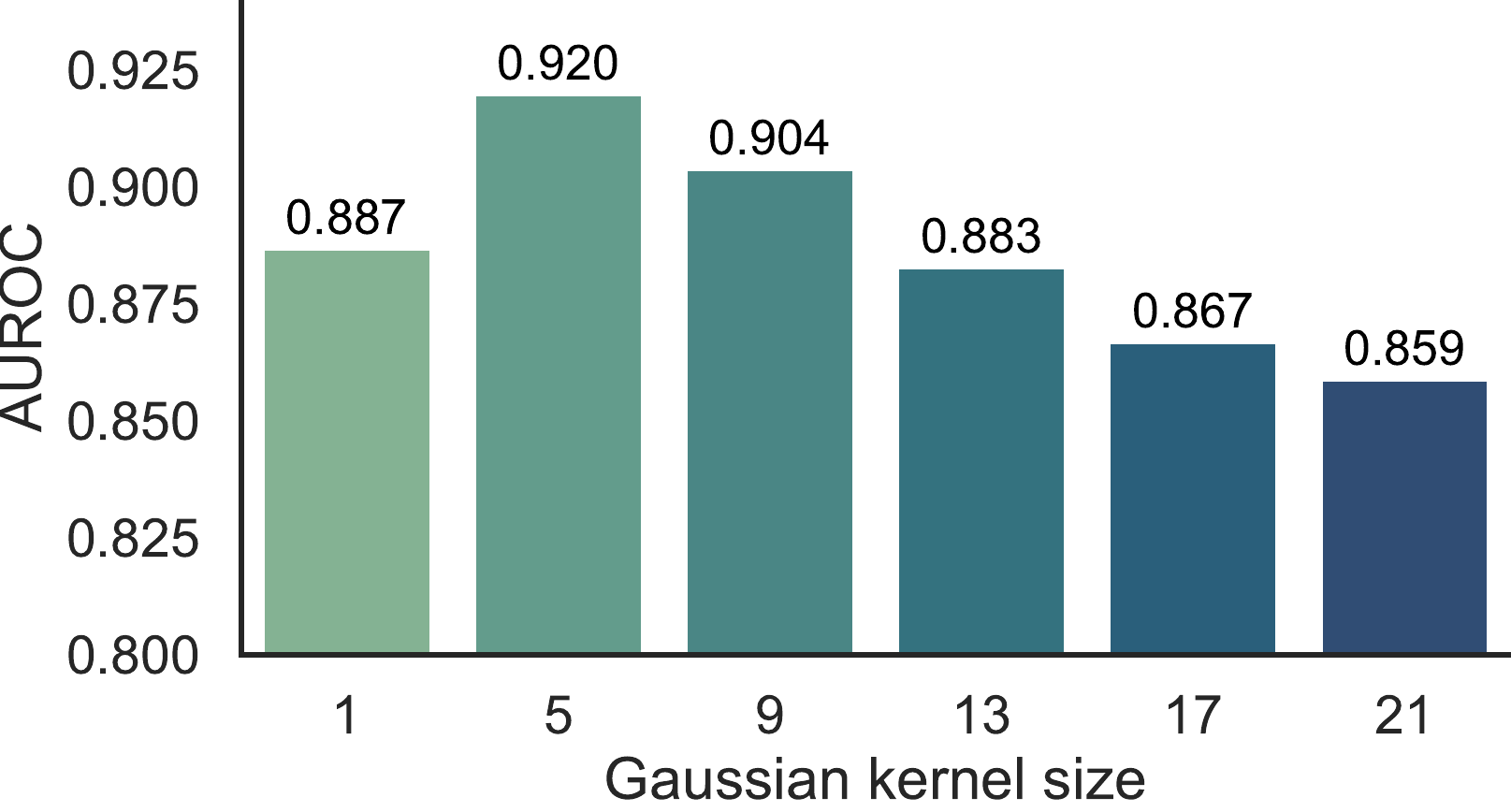}  
      \caption{PixelCNN++}
      \label{fig:sub-second2}
    \end{subfigure}
    \caption{Ablation on Gaussian kernel sizes in GLOW and PixelCNN++. CIFAR-10 is the in-distribution data. Results are averaged over all OOD test datasets.} %
    \label{fig:glow_ablation_kernel}
    \end{figure}

Similar to the ablation studies on Gaussian kernels in  VAE, we train GLOW~\cite{kingma2018glow} and PixelCNN++~\cite{salimans2017pixelcnn++} models with different kernel
sizes on CIFAR-10, and evaluate the OOD detection
performance respectively. The average AUROC across all
OOD datasets is shown in Figure ~\ref{fig:glow_ablation_kernel}. Results also suggest that \method{} is not sensitive to the choice of kernel size.

 \section{Model Overview for PixelCNN++}
\begin{figure}[htbp] %
	\begin{center}
    \includegraphics[width=0.4\linewidth]{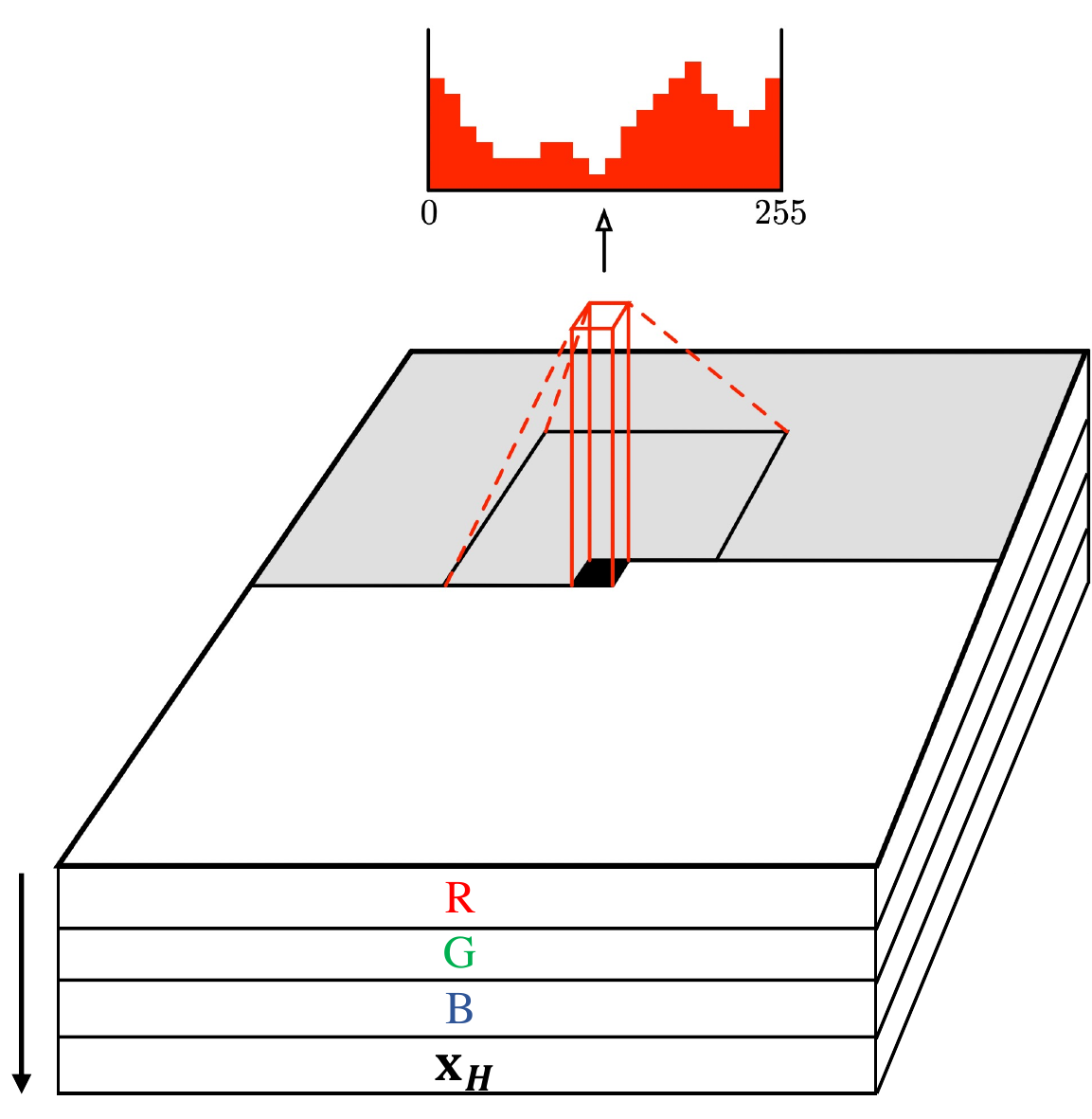} %
	\end{center}
	   \caption{Overview of PixelCNN++ under \method{}.}
	\label{fig:pixelcnn_overview}
 \end{figure}

The overview of PixelCNN++~\cite{salimans2017pixelcnn++} under \method{} is shown in Figure~\ref{fig:pixelcnn_overview}. For each pixel, four
channels ($\mathbf{x}_H$, R, G, B) are modeled successively, with $\mathbf{x}_H$ conditioned on (R, G, B),  B  conditioned on (R, G), and G conditioned on R. Here $\mathbf{x}_H$ denotes the high-frequency features given input $\mathbf{x}$ in \method{}. The sequential prediction is achieved by splitting the feature maps at every layer of the network into four and adjusting the centre values of the mask tensors. The 256 possible values for each channel are then modeled using the softmax.

PixelCNN++~\cite{salimans2017pixelcnn++} consists of a stack of masked convolution layers that takes an $N \times N \times 4$ image
as input and produces $N \times N \times 4 \times 256$  predictions as output. The use of convolutions allows the
predictions for all the pixels to be made in parallel during training. During sampling the predictions are sequential: every time a pixel is predicted, it is fed back into the network to predict the next pixel.

\section{Model Overview for GLOW}
\begin{figure}[htbp] %
	\begin{center}
    \includegraphics[width=\linewidth]{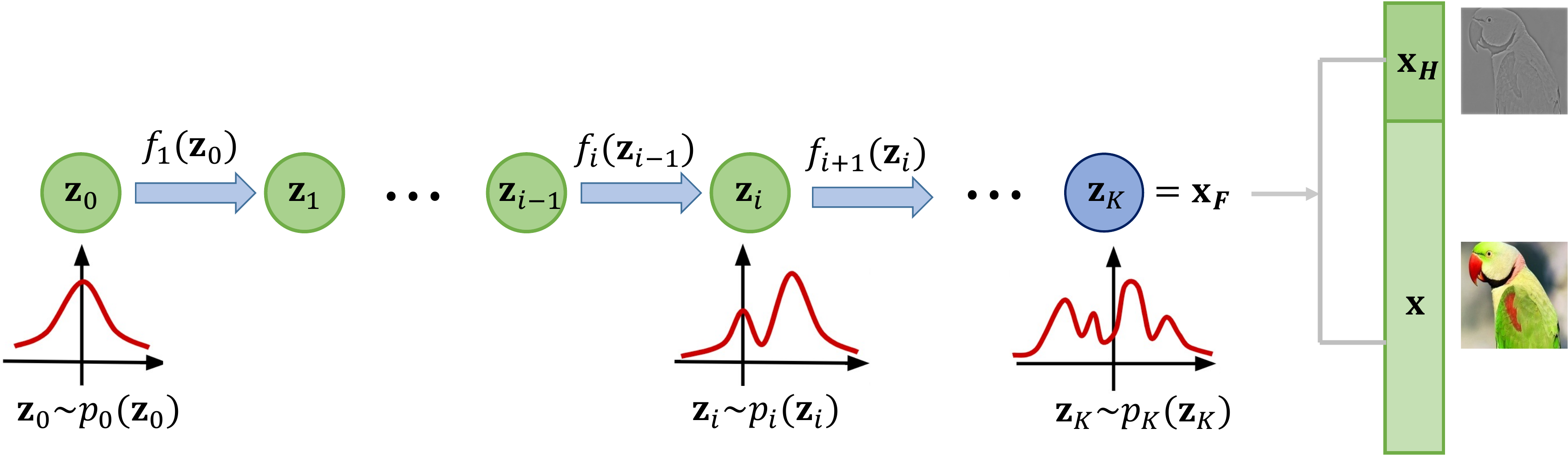} %
	\end{center}
	   \caption{Overview of the GLOW  under \method{}.}
	\label{fig:glow_overview}
 \end{figure}

The overview of GLOW~\cite{kingma2018glow} under \method{} is shown in Figure~\ref{fig:glow_overview}.

 \begin{figure*}[htbp]
	\begin{center}
    \includegraphics[width=\linewidth]{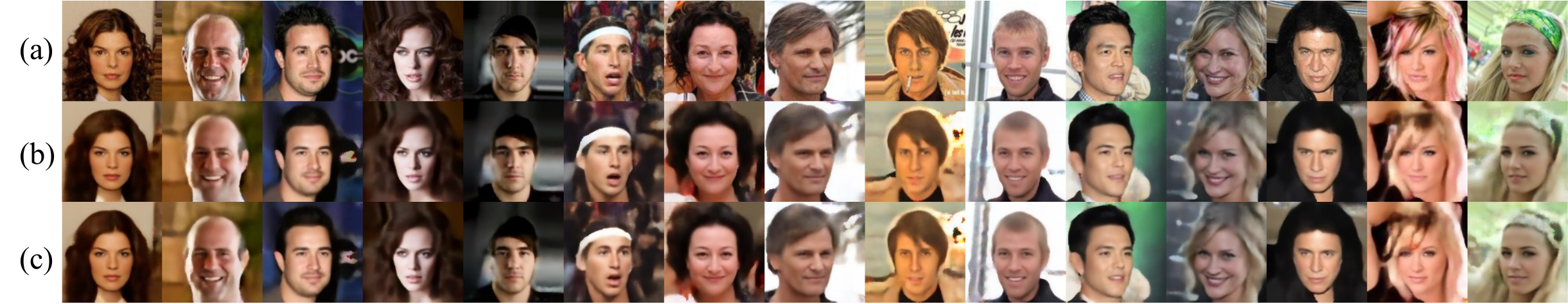} %
	\end{center}
	   \caption{The qualitative reconstruction results for CelebA under VAE. (a) shows the source images randomly picked from CelebA. (b) Shows the reconstruction results under the vanilla VAE training scheme. (c) shows the results when VAE is trained jointly with pixel level input and the high frequency information.}
	\label{fig:recon}
 \end{figure*}

GLOW consists of a series of steps of flow, combined in a multi-scale architecture~\cite{kingma2018glow}.
Each step of flow consists of actnorm followed by an invertible $1 \times 1$ convolution and a coupling layer. The GLOW  model has a depth of flow $K$, which maps the latent variable $\mathbf{z}_0$ to input $\mathbf{x}_F$.
 
\section{More Training Details}

We provide more details for training on each of the architecture: VAE, GLOW and PixelCNN+.

 (1) We train VAE using  learning rate $1 \times 10^{-3}$ and Adam optimizer~\cite{kingma2014adam}. 
The learning rate is decayed by half every 30 epochs. The encoder of VAEs consists of four convolutional layers of kernel size 4, strides 2 without biases.
The decoder of VAEs has a symmetric structure to the encoder, reconstructing the image pixel-wise.
In VAE, the distribution of the  latent code prior $p(\mathbf{z})$ and the variational posterior $q_{\phi}(\mathbf{z}|\mathbf{x})$ are set to be Gaussians. 
In our experiments, the dimension for the latent code $p(\mathbf{z})$ is set to be 200 for CIFAR-10 and CelebA,  and 100 for Fashion-MNIST. 

(2) For GLOW, the number of the hidden channels is 400 for CIFAR-10 and 200 for Fashion-MNIST. 
3-layer networks are used in the coupling blocks. We use two blocks of 16 flows for Fashion-MNIST, and three blocks of 8 flows with multi-scale for CIFAR-10. 

(3) In PixelCNN++, all residual layers use 192 feature maps and a dropout rate of 0.5.

\section{Qualitative Results of Image Generation}

\method{} not only significantly improves the  OOD detection performance, but also preserves the generative capability. Figures~\ref{fig:recon} shows the visualizations of the reconstruction results under the vanilla VAE and \method{} for CelebA, where the reconstruction quality of \method{} is not compromised. We further  quantitatively measure the reconstruction results in Table~\ref{tab:table_reconstruct}. \method{} achieves stronger results measured by both pixel-level and perception level metrics, including mean-square error (MSE), mean absolute error (MAE), peak signal-to-noise ratio (PSNR),  and SSIM.

 \begin{table}[htbp]%
  \begin{center}
    \caption{CelebA reconstruction performance, measured by the image reconstruction quality  between vanilla VAE and \method{}. Evaluation metrics includes mean-square error~(MSE), mean absolute error~(MAE), peak signal-to-noise ratio~(PSNR)~\cite{de2003improved}, and SSIM~\cite{wang2004image}. $\uparrow$ means that higher value represents better image quality, and vice versa.  
  }
  \label{tab:table_reconstruct}
  \small
  \begin{tabular}{ l|c|c}
      \toprule
        \diagbox{Metrics}{Method}~&~vanilla VAE~&~\method{} \\
      \hline
        MSE $\downarrow$~&~0.0041 &\textbf{0.0037} \\
        MAE $\downarrow$~&~0.0352 &\textbf{0.0337} \\
        PSNR $\uparrow$~&~24.657 &\textbf{25.097} \\
        SSIM $\uparrow$~&~0.7662 &\textbf{0.7806} \\
       \bottomrule
  \end{tabular}
  \end{center}
  
\end{table}

\section{Social Impact}
This paper aims to improve the reliability and safety of modern neural networks, particularly generative model families. Our study can lead to direct benefits and societal impacts when deploying machine learning models in the real world. Our work does not involve any human subjects or violation of legal compliance. We do not anticipate any potentially harmful consequences. Through our study and releasing our code, we hope to raise stronger research and societal awareness towards the problem of out-of-distribution detection.

\end{document}